\documentclass[sigconf, nonacm]{acmart}

\AtBeginDocument{%
    }

\copyrightyear{2025}
\acmYear{2025}
\acmConference[SC Workshops '25]{Workshops of the International Conference for High Performance Computing, Networking, Storage and Analysis}{November 16--21, 2025}{St Louis, MO, USA}
\acmBooktitle{Workshops of the International Conference for High Performance Computing, Networking, Storage and Analysis (SC Workshops '25), November 16--21, 2025, St Louis, MO, USA}
\acmDOI{10.1145/3731599.3767340}
\acmISBN{979-8-4007-1871-7/2025/11}

\newif\ifredact
\redactfalse % disable redaction

\usepackage{array}
\usepackage{environ}
\usepackage{balance}
\usepackage{xcolor}
\usepackage{float}
\usepackage{subfigure} % or subcaption
\floatstyle{ruled}
\usepackage[nolist]{acronym}

\newcommand{\redact}[1]{%
\ifredact
    \colorbox{black}{\phantom{#1}}%
\else
    #1%
\fi
}

\begin{document}

% Define acronyms
\begin{acronym}
  \acro{HPC}{high-performance computing}
  \acro{AI}{artificial intelligence}
  \acro{PDF}{probability density function}
  \acro{DNS}{direct numerical simulation}
  \acroplural{DNS}[DNS]{direct numerical simulations}
  \acro{CFD}{computational fluid dynamics}
  \acro{SOTA}{state-of-the-art}
  \acro{LLM}{large language models}
\end{acronym}


\title[Intelligent Sampling of Extreme-Scale Turbulence Datasets]{Intelligent Sampling of Extreme-Scale Turbulence Datasets for Accurate and Efficient Spatiotemporal Model Training}

%%
%% The abstract is a short summary of the work to be presented in the article.
\begin{abstract}
With the end of Moore's law and Dennard scaling, efficient training increasingly requires rethinking data volume. Can we train better models with significantly less data via intelligent subsampling? To explore this, we develop SICKLE, a sparse intelligent curation framework for efficient learning, featuring a novel maximum entropy (MaxEnt) sampling approach, scalable training, and energy benchmarking. We compare MaxEnt with random and phase-space sampling on large direct numerical simulation (DNS) datasets of turbulence. 
Evaluating SICKLE at scale on Frontier, we show that subsampling as a preprocessing step can, in many cases, improve model accuracy and substantially lower energy consumption, with observed reductions of up to 38×.
\end{abstract}

% <concept_id>10010405.10010432.10010442</concept_id>
% <concept_desc>Applied computing~Mathematics and statistics</concept_desc>


\author{Wesley Brewer}
\affiliation{%
  \institution{Oak Ridge National Laboratory}
  \city{Oak Ridge}
  \country{USA}
}
\email{brewerwh@ornl.gov}

\author{Murali Meena Gopalakrishnan}
\affiliation{%
  \institution{Oak Ridge National Laboratory}
  \city{Oak Ridge}
  \country{USA}
}
\email{gopalakrishm@ornl.gov}

\author{Matthias Maiterth}
\affiliation{%
  \institution{Oak Ridge National Laboratory}
  \city{Oak Ridge}
  \country{USA}
}
\email{maiterthm@ornl.gov}

\author{Aditya Kashi}
\affiliation{%
  \institution{Oak Ridge National Laboratory}
  \city{Oak Ridge}
  \country{USA}
}
\email{kashia@ornl.gov}

\author{Jong Youl Choi}
\affiliation{%
  \institution{Oak Ridge National Laboratory}
  \city{Oak Ridge}
  \country{USA}
}
\email{choij@ornl.gov}

\author{Pei Zhang}
\affiliation{%
  \institution{Oak Ridge National Laboratory}
  \city{Oak Ridge}
  \country{USA}
}
\email{zhangp1@ornl.gov}

\author{Stephen Nichols}
\affiliation{%
  \institution{Oak Ridge National Laboratory}
  \city{Oak Ridge}
  \country{USA}
}
\email{nicholsss@ornl.gov}

\author{Riccardo Balin}
\affiliation{%
  \institution{Argonne National Laboratory}
  \city{Lemont}
  \country{USA}
}
\email{rbalin@anl.gov}

\author{Miles Couchman}
\affiliation{%
  \institution{York University}
  \city{Toronto}
  \country{Canada}
}
\email{mmpc@yorku.ca}

\author{Stephen de Bruyn Kops}
\affiliation{%
  \institution{University of Massachusetts Amherst}
  \city{Amherst}
  \country{USA}
}
\email{debk@umass.edu}

\author{P.K. Yeung}
\affiliation{%
  \institution{Georgia Institute of Technology}
  \city{Atlanta}
  \country{USA}
}
\email{pk.yeung@ae.gatech.edu}

\author{Daniel Dotson}
\affiliation{%
  \institution{Georgia Institute of Technology}
  \city{Atlanta}
  \country{USA}
}
\email{ddotson6@gatech.edu}

\author{Rohini Uma-Vaideswaran}
\affiliation{%
  \institution{Georgia Institute of Technology}
  \city{Atlanta}
  \country{USA}
}
\email{rvaideswaran3@gatech.edu}

\author{Sarp Oral}
\affiliation{%
  \institution{Oak Ridge National Laboratory}
  \city{Oak Ridge}
  \country{USA}
}
\email{oralhs@ornl.gov}

\author{Feiyi Wang}
\affiliation{%
  \institution{Oak Ridge National Laboratory}
  \city{Oak Ridge}
  \country{USA}
}
\email{fwang2@ornl.gov}

\renewcommand{\shortauthors}{\redact{Brewer} et al.}



% \received{14 April 2025}
% \received[revised]{14 July 2025}
% \received[accepted]{25 August 2025}

\maketitle
    
\section{Introduction}

\begin{figure}
\centering
\includegraphics[width=1\linewidth]{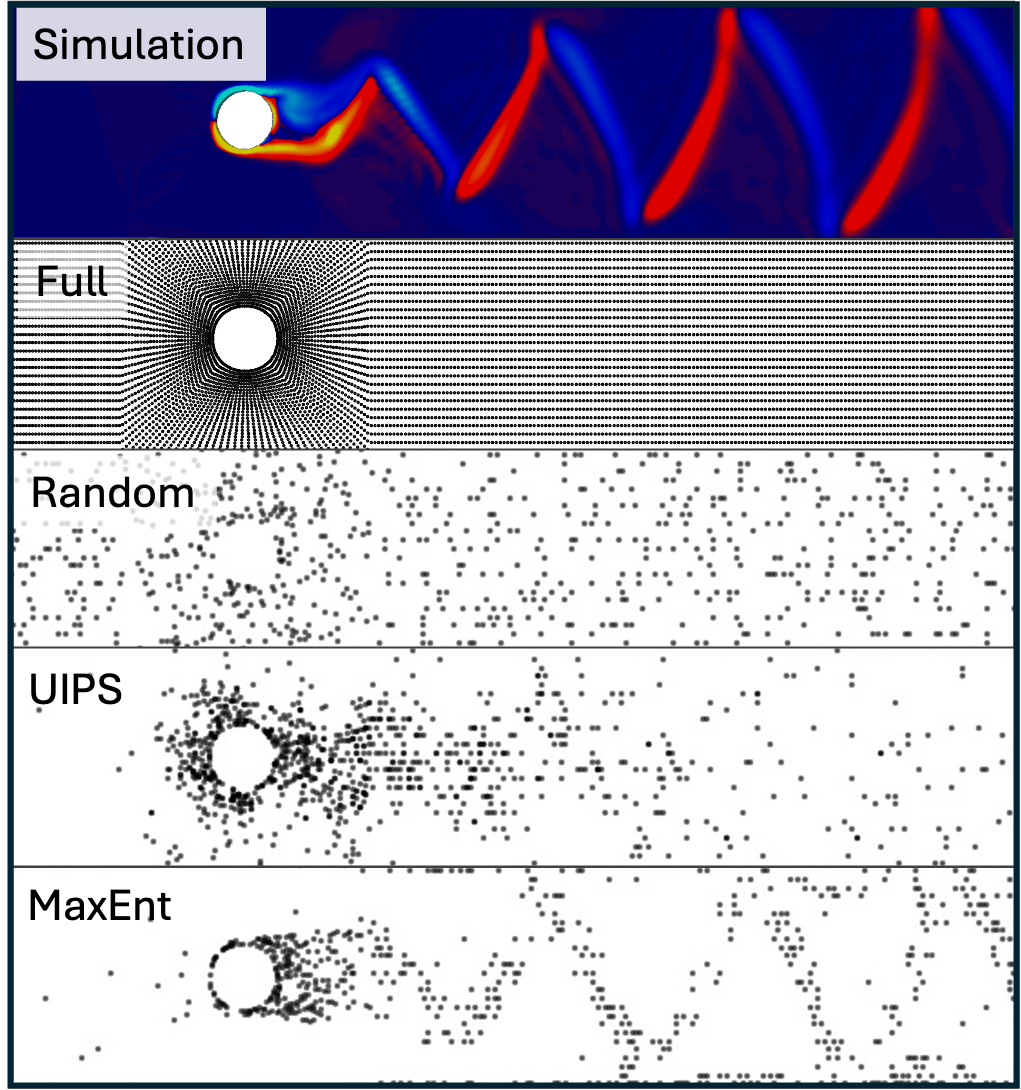}
\caption{Demonstrating different subsampling strategies for 2D flow over cylinder case (OF2D), 10\% sampling rate.}
\label{fig:cylinder}
\end{figure}

Recent successes with foundation models in natural language processing and computer vision have sparked significant interest in their development for scientific and engineering applications \cite{bommasani2021opportunities, kirillov2023segment, hu2023scaling, nguyen2023climax}. 
Training large-scale foundation models typically requires extensive data to achieve high accuracy and broad applicability; the costs associated with training \ac{SOTA} \ac{LLM} are rapidly approaching billions of dollars \cite{cottier2024rising}. 
Reducing data volume through intelligent selection is key to making large-scale model training more tractable and sustainable.
Figure~\ref{fig:cylinder} illustrates this concept with a 2D cylinder flow example, showing how different subsampling strategies capture distinct flow features even at the same sampling rate.

\begin{figure*}[t]
    \centering
    \includegraphics[width=0.65\linewidth]{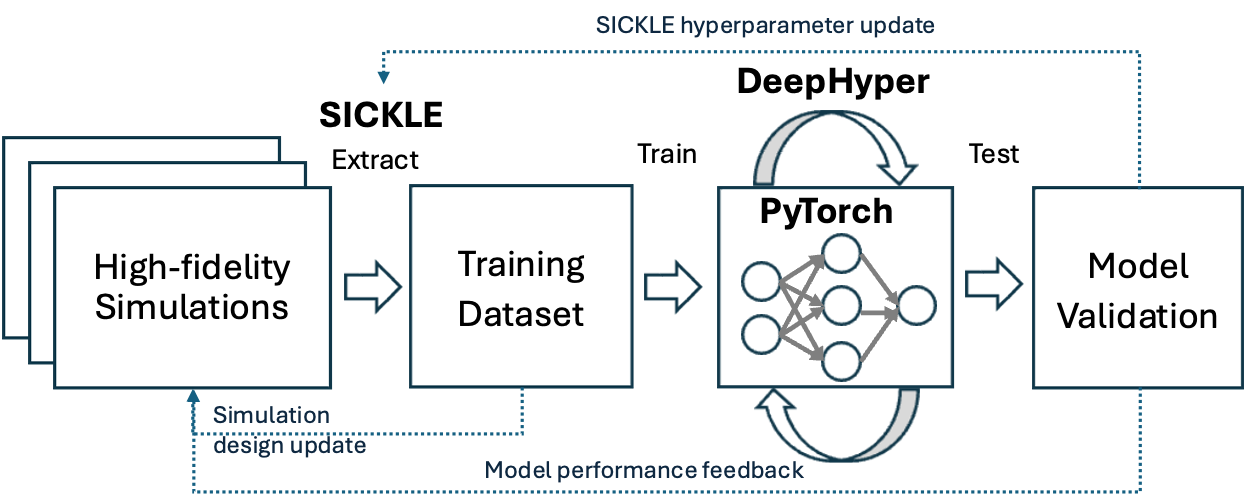}
    \caption{Spatiotemporal model training workflow.}
    \label{fig:workflow}
\end{figure*}

It has often been said that data is the lifeblood of AI. Data movement, typically, is the most energy-intensive aspect of high-performance computing (HPC); for example, the energy needed to move a double-precision datum across a system is over 100 times greater than computing it~\cite{kogge2013exascale, peterka2024priority}. Byna et al.~\cite{byna2024data} suggest that future gains in energy efficiency must come from reducing data movement, which can be achieved through advanced artificial intelligence (AI)-based data reduction methods that minimize unnecessary data transfers between workflow stages. Because all data points are not equally informative, significant volumes of data can often be considered redundant or less valuable~\cite{byna2024data}. Thus, intelligently extracting the most valuable data samples becomes crucial for efficient dataset curation.

% using the MATEY framework—
To demonstrate our approach, we focus on the challenging use case of training turbulence models, a representative example of complex scientific datasets that differ significantly from typical text or image-based AI workloads. Turbulence is an extreme multiscale, chaotic, and nonlinear physical phenomenon that plays a critical role in many applications. Capturing its physics accurately requires high-fidelity \acp{DNS}, which are computationally demanding and can generate petabytes of data. These large datasets pose both storage and processing challenges, necessitating scalable analysis methods. In this work, we focus specifically on curating such large \ac{DNS} datasets for the purpose of training machine-learned surrogates through intelligent sampling.

A central goal of turbulence research is to develop predictive models that generalize across flow regimes and scales, and ultimately help understand and predict turbulent behavior. A promising step toward this vision is the development of foundation models for turbulence, which are currently being explored in this and related work \cite{zhang2024matey}. However, training such models is challenging—they require vast and diverse datasets, often petabytes in size, generated from high-fidelity simulations or experiments. While techniques such as compression, coarse-graining, and data reduction have been considered, e.g., \cite{fukami2019super, rowley2009spectral, carlberg2011efficient}, this paper focuses on intelligent sampling as a scalable and effective strategy for dataset curation.

The training and deployment workflow—illustrated in Fig.~\ref{fig:workflow}—involves running large-scale simulations, extracting representative training data, tuning the neural network architecture, and validating the model before deployment. Depending on the use case, the model can serve as a standalone surrogate or augment lower-fidelity simulations. While this workflow is tailored to turbulence, it shares core components with many AI-coupled HPC workflows~\cite{brewer2024ai}, making it a representative testbed for evaluating sparse learning frameworks like SICKLE.

In this paper, we explore training data sparsity as it relates to energy efficiency. 
\textit{We hypothesize that a maximum entropy-based subsampling approach can, in many cases, yield more accurate models with improved computational efficiency compared to training on the full dataset, thereby also reducing energy consumption.}
In order to test our hypothesis, we explore sparsification methods on an exascale supercomputer for the case of one of the most computationally expensive simulations, \ac{DNS} of stratified and isotropic turbulence. The outcome of this study is a framework called SICKLE -- Sparse Intelligent Curation frameworK for Learning Efficiently -- designed to enable machine learning (ML) on intelligently extracted data subsets. SICKLE includes several \ac{SOTA} approaches for intelligently curating subsets of training data, including: random sampling, Latin hypercube sampling (LHS), stratified sampling, uniform-in-phase-space selection~\cite{hassanaly2023uniform}, and a scalable maximum-entropy-weighted stratified sampling approach (MaxEnt). Our framework also includes methods for benchmarking performance, accuracy, and energy efficiency. Moreover, our framework provides a convenient way to significantly reduce file storage requirements, by storing feature-rich subsampled datasets, as well as providing means of visualizing them. To support reproducibility and adoption, SICKLE is available via GitHub.\footnote{\url{https://github.com/at-aaims/sickle}}

Our paper is structured as follows: in Section \ref{sec:background}, we give a background of other related research; in Section \ref{sec:datasets}, we introduce the datasets used in the study; in Section \ref{sec:sampling} we introduce two different \ac{SOTA} sampling strategies and evaluate them on the simulation datasets; in Section \ref{sec:training}, we use the subsampled datasets to train models for turbulence and discuss the results; in Section \ref{sec:benchmarking} we explore the efficiency and scalability of the subsampling methods. Finally, in Section \ref{sec:conclusions} we discuss conclusions and future work. 
%Throughout the various sections, we have highlighted valuable lessons learned, denoted as ``findings''.

\section{Background}
\label{sec:background}

Historically, advances in HPC have relied on hardware improvements like transistor scaling (Moore's law~\cite{moore1998cramming}) and voltage scaling (Dennard scaling~\cite{dennard1974design}). However, the end of these trends has made further gains in computational capability and energy efficiency increasingly challenging~\cite{shalf2020future}. A major bottleneck in exascale computing is the substantial energy consumption associated with data movement rather than computation itself~\cite{kestor2013quantifying}. Although strategies like reduced-precision arithmetic and hardware innovations have been explored, these alone are insufficient for post-exascale computing. Intelligent data reduction at the source via subsampling presents an alternative solution.

While massive datasets in HPC and scientific computing often assume that more data inherently leads to higher accuracy, evidence suggests that intelligently selecting a curated subset of informative samples can yield more accurate and efficient models. For example, Zhao et al.~\cite{zhao2022adaptive} note that ``lacking systematic sampling algorithms for dynamics learning may lead to less accurate models.'' Redundant or low-variance data can dilute training signals and lead to overfitting, especially in scientific datasets like turbulence. This phenomenon has also been observed more broadly: Toneva et al.~\cite{toneva2019forgetting} show that redundant ``easy'' examples contribute little to learning, while Xia et al.~\cite{xia2024rethinking} demonstrate that large-scale language models can train more effectively on carefully selected subsets than on the full dataset. Wang et al.~\cite{wang2023deep} mention that ``there is a burgeoning interest in exploring effective subsampling techniques,'' pointing to the growing recognition of these issues.

Intelligent subsampling has been applied across diverse scientific domains—including aerospace and astrophysics~\cite{wissink2025multi,biswas2020probabilistic,taghribi2022asap}, biology~\cite{oues2023mdsubsampler,kleiman2023adaptive,wang2023deep}, climate and space weather~\cite{leuridan2023polytope,lavasa2021assessing,biswas2020probabilistic}, combustion~\cite{hassanaly2023uniform,biswas2020probabilistic}, and materials science~\cite{robinson2023towards,becker2022unsupervised,karabin2020entropy}. Other applications span complex engineering dynamics~\cite{zhao2022adaptive,mostafa2022machine,lin2018intelligent}, remote sensing~\cite{verrelst2020intelligent,zhang2022machine}, and large language modeling~\cite{chen2021fedgraph,yin2024entropy}. Despite this, widespread adoption remains limited, often due to narrow contexts and lack of systematic evaluation, especially regarding energy efficiency at scale.

Subsampling methods generally fall into five categories:

1. \textbf{Random and Quasi-Random Sampling.} Includes uniform random sampling and Latin hypercube sampling, which offers simplicity and unbiased coverage~\cite{yuan2020evaluation, raissi2019physics}.

2. \textbf{Stratified and Clustering-Based Sampling.} Data is partitioned into strata, selecting representative samples from each.

3. \textbf{Projection-Based Methods.} Techniques like proper orthogonal decomposition and dynamic mode decomposition project data onto low-rank subspaces~\cite{berkooz_pod_1993, schmid_dmd_2010}. 

4. \textbf{Physics-informed and Specialized Sampling.} Domain-specific knowledge guides sample selection, such as uniform-in-phase-space methods~\cite{hassanaly2023uniform}.

5. \textbf{Model-Driven and Adaptive Sampling.} Methods adaptively select informative samples based on criteria like uncertainty and generalization error~\cite{rotskoff2022active, wight2020solving, zhao2022adaptive, o2024quantifying}.

Our work uniquely integrates maximum entropy principles into an automated and scalable subsampling framework for training large-scale foundation models. While entropy-based sampling has theoretical roots in information theory~\cite{shannon1948mathematical, jaynes1957information, shewry1987maximum}, we extend and scale this concept to high-dimensional, terabyte-scale 3D turbulence datasets within our comprehensive SICKLE framework, including benchmarking and energy efficiency evaluations.

Several critical gaps remain in current research on intelligent subsampling: (1) Studies rarely quantify the impact of subsampling on computational efficiency, (2) such methods are rarely demonstrated for large datasets (terabytes to petabytes), and (3) few studies systematically assess subsampling's effects on model accuracy and generalization.
Our work addresses these gaps through:

\begin{enumerate}
\item Innovation and development of an information-theoretic, scalable framework for analyzing CFD datasets, based on the principle of maximum entropy. The framework, called SICKLE, also features a pluggable architecture that makes it easy to integrate other sampling strategies.
\item Demonstration of the method across multiple DNS datasets with comparison to baseline sampling methods (e.g., random sampling) and state-of-the-art methods (e.g., phase-space sampling).
\item Systematic benchmarking on the Frontier supercomputer, evaluating energy, scalability, computational performance, and accuracy.
\item Demonstrating intelligent subsampling significantly reduces energy consumption while maintaining or enhancing accuracy compared to full dataset training.
\end{enumerate}

\renewcommand{\arraystretch}{1.5}
\begin{table*}[t]
    \centering
    \caption{Summary of datasets used in study with K-means cluster variable (KCV), neural network input/output parameters. Here, $C$ and $\widetilde{C^{''2}}$ are respectively the progress variable and its filtered variance, $D$ is drag, $p$ is pressure, $q$ represents potential vorticity, $\rho$ is density, $u$, $v$, $w$ are velocity components, $\epsilon$ is dissipation rate, and $\Omega$ denotes enstrophy. Time column is number of available time steps. T-G[i] stands for Taylor Green initialization.}
    \begin{tabular}{ll|cc|cc|ccc}
        \hline
        \textbf{Label} & \textbf{Description} & \textbf{Space (grid points)} & \textbf{Time} & \textbf{Size} & \textbf{KCV} & \textbf{Input} & \textbf{Output} \\
        \hline
        TC2D & 2D Turbulent Combustion~\cite{hassanaly2023uniform} & 400k & 1 & 31MB & -- & $C$, $\widetilde{C^{''2}}$ & -- \\
        OF2D & 2D Laminar Flow Over Cylinder~\cite{brewer2023entropy} & 10800 & 100 & 300MB & $p$ & $u, v$ & $D$ \\
            \hline
            SST-P1F4  & 3D T-G[i] time evolving Pr=1~\cite{riley2003dynamics} & 512×512×256 & 125 & 376GB & $q$ & $u, v, w$ & $p$ \\
            % Green & 3D Homogeneous Isotropic & 12288x6144x12288 & 6 & 72TB & Reconstruction \\
        SST-P1F100 & 3D Forced stratified turbulence \cite{portwood2016robust} & 4096×1024×4096 & 10 & 5TB & $\rho$ & $u, v, w, \rho$ & $\epsilon$ \\
        % SST-PiF50 & 3D Homogeneous time evolving Pr=50 & 37632×4704×37632 & 2 & 200TB & $q$ & $u, v, w$, $\rho$ & $p$ \\
        % SST-P50 & 3D T-G[i] time evolving Pr=50~\cite{petropoulos2025modelling} & 3584×3584×1792 & 1681 & 414TB & $q$ & $u, v, w$, $\rho$ & $p$ \\
            \hline
            GESTS-2048 & 3D Forced isotropic turbulence~\cite{yeung2025gpu} & 2048x2048x2048 & 1 & 188GB & $\Omega$ & $u, v, w$, $\epsilon$ & $p$ \\
            GESTS-8192 & 3D Forced isotropic turbulence~\cite{yeung2025gpu} & 8192x8192x8192 & 1 & 12TB & $\Omega$ & $u, v, w$, $\epsilon$ & $p$ \\
            % GESTS-32768 & & 32768×32768×32768 & 1 & 768TB & $\Omega$ & $u, v, w$, $\epsilon$ & $p$ \\
        \hline
    \end{tabular}
    \label{tab:datasets}
\end{table*}

\section{Simulation Datasets}
\label{sec:datasets}

To evaluate our hypothesis, we use several \ac{CFD} simulation datasets ranging from small 2D cases to large-scale 3D simulations (Table~\ref{tab:datasets}). Initial development focused on lightweight 2D datasets to iterate and test core functionality. The framework was subsequently extended and parallelized for large 3D turbulence datasets.\footnote{Datasets such as TC2D~\cite{hassanaly2023uniform} are publicly available at \url{https://github.com/NREL/Phase-space-sampling}. Additional datasets will be made publicly available upon publication.}

\paragraph{2D Simplified Datasets (TC2D/OF2D)}
We initially developed and tested phase-space and MaxEnt sampling on two simplified cases: a downsampled 2D turbulent combustion dataset~\cite{hassanaly2023uniform,yellapantula2021deep}, and a 2D laminar flow over a cylinder case~\cite{brewer2023entropy}. The cylinder flow simulation was conducted using OpenFOAM~\cite{jasak2007openfoam} at a Reynolds number of 1267, employing a multi-block structured grid body-fitted to the cylinder with no-slip conditions.

\paragraph{3D Stably Stratified Turbulence (SST)}
De Bruyn Kops and collaborators have produced an ensemble of DNS probing stably stratified turbulence influenced by buoyancy~\cite{riley2003dynamics,portwood2016robust,petropoulos2025modelling}. These simulations involve an array of Taylor-Green vortices transitioning from laminar to turbulent states, eventually re-laminarizing under stabilizing buoyancy forces. The largest dataset includes four flow variables (density and velocity components) across 1681 snapshots, totaling 414 terabytes. 
Intelligent subsampling methods are essential for distilling such extensive and anisotropic datasets for efficient ML training.

\paragraph{3D Forced Isotropic Turbulence (GESTS)}
Yeung et al.~\cite{yeung2025gpu} conducted the largest \ac{DNS} simulation to date using the GPU-based code suite GESTS on 8192 nodes of Frontier, achieving resolutions up to 32768$^3$. Their simulations utilize Fourier pseudo-spectral methods with GPU-aware MPI and OpenMP offloading. Nonlinear terms in the Navier-Stokes equations are handled in physical space, while differentiation and time evolution occur in wavenumber space. Solution checkpoints are stored in wavenumber space, with a post-processing step converting data into 1024$^3$ bricks in physical space for evaluation.

\section{Intelligent Sampling}
\label{sec:sampling}

\begin{figure}
    \centering
    \includegraphics[width=0.62\linewidth]{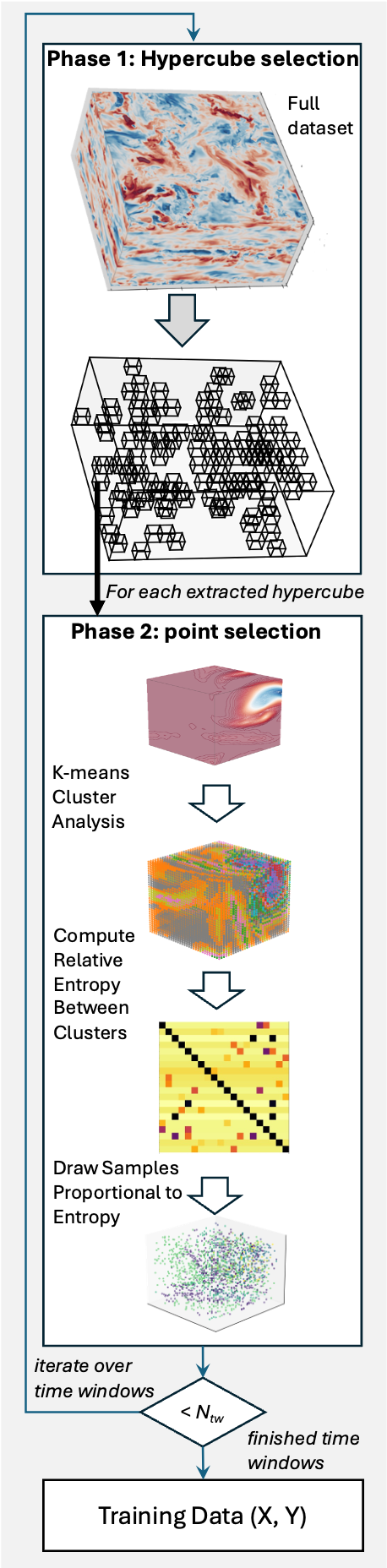}
    \caption{MaxEnt hypercube selector and point sampler.} 
    \label{fig:methods}
\end{figure}

Numerous approaches to data sub-sampling exist, including random, stratified, Latin hypercube, and importance sampling. We focus specifically on stratified maximum entropy (MaxEnt) sampling, illustrated in Fig.~\ref{fig:methods}. This two-phase method initially selects $N$ hypercubes either randomly or by entropy and then selects subsamples using random, MaxEnt, or phase-space sampling. If structured cubes are required by neural networks, the second step is skipped. 
We note that training on the entire DNS dataset (all grid points and timesteps) is computationally infeasible. Accordingly, `full' in our comparisons refers to fully sampled hypercubes of size 32³, which serve as the densest feasible baseline under our workflow.
Fig.~\ref{fig:cylinder} illustrates the different subsampling methods, with MaxEnt more effectively capturing the wake flow features compared to the other approaches.

% \begin{figure}
%     \centering
%     \includegraphics[width=0.8\linewidth]{figs/OF2D.png}
%     \caption{OF2D case demonstrating different subsampling strategies for 10\% sampling of full dataset.}
%     \label{fig:cylinder}
% \end{figure}

\subsection{Max-entropy sampling}

\paragraph{Phase 1: MaxEnt hypercube selection (Hmaxent)}
Phase one reduces dense datasets into sparse hypercubes via clustering and entropy-based subsampling. The procedure involves clustering, computing probability distributions, determining entropy and node strengths, and entropy-weighted random sampling. We utilize MPI for parallelization and MiniBatchKMeans for efficient clustering.

\paragraph{Computation of Entropy}
Entropy is computed using the Kullback-Leibler divergence:

\begin{equation}
D = \sum p \log(p / q)
\end{equation}

An adjacency matrix $\boldsymbol{A}$ captures relative entropies between clusters, with node strengths being the row-level sum:

\begin{equation}
A_{ij} = \sum \boldsymbol{P}(C_i) \log\left(\boldsymbol{P}(C_i) / \boldsymbol{P}(C_j)\right)
\label{eq:adjacency}
\end{equation}

\paragraph{Phase 2: MaxEnt point selection (Xmaxent)}
The second step involves clustering and entropy selection within each hypercube. Simulation data are clustered on a target variable, distributions computed, and samples drawn based on node strengths.

\subsection{Phase-space sampling (UIPS)}
\begin{figure}
    \centering
    \includegraphics[width=3.3in]{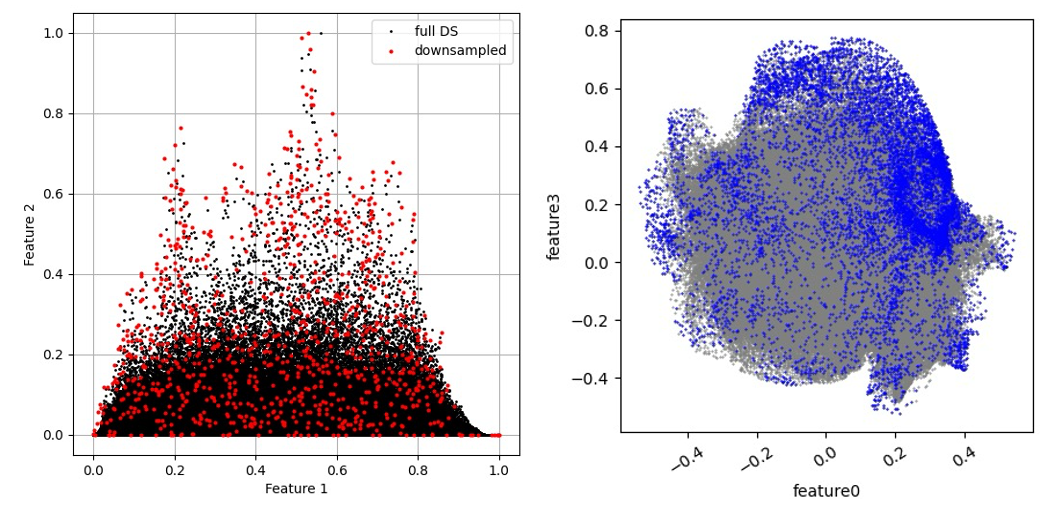}
    \caption{UIPS works well for 2D turbulent combustion dataset (left), however it exhibits clumping behavior when applied to the 3D anisotropic SST-P1F4 dataset (right). Features correspond to the inputs listed in Table \ref{tab:datasets}.}
    \label{fig:uips}
\end{figure}

Following \cite{hassanaly2023uniform}, phase-space sampling involves generating \acp{PDF} either through binning or normalizing flows. Although normalizing flows handle high-dimensional spaces well, binning was adopted for temporal dimensions due to implementation simplicity. 
Fig.~\ref{fig:uips} illustrates that UIPS performs well on 2D datasets but tends to concentrate samples unevenly in 3D anisotropic flows, where uniformity breaks down due to the underlying data structure.

\begin{figure*}
    \centering
    \subfigure[OF2D]{\includegraphics[width=0.33\linewidth]{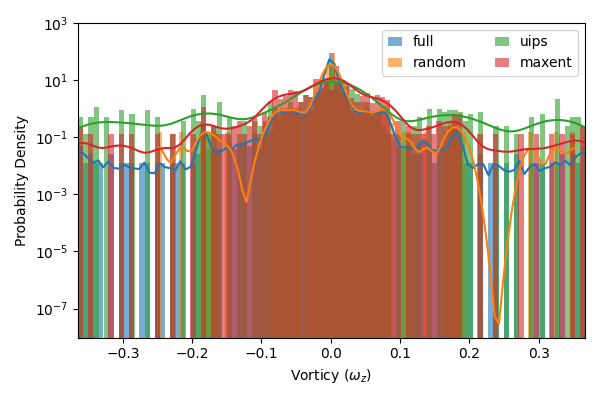}} 
    \subfigure[SST-P1F4]{\includegraphics[width=0.33\linewidth]{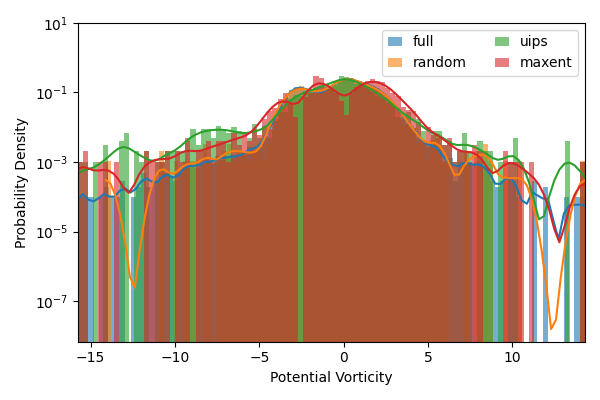}}
    \subfigure[GESTS-2048]{\includegraphics[width=0.33\linewidth]{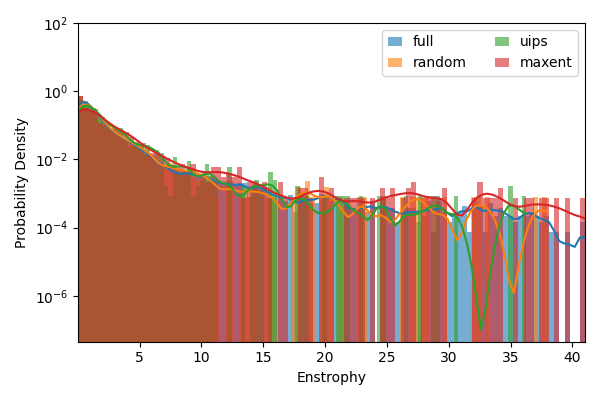}}
    \caption{\acp{PDF} of subsampling methods (10\% sampling of full dataset).}
    \label{fig:pdf}
\end{figure*}

Fig.~\ref{fig:pdf} demonstrates MaxEnt’s effectiveness in covering tail regions that contribute to improved model generalization, especially as compared to random baseline.

\subsection{Temporal Sampling}
\label{sec:temporal_sampling}

In addition to spatial sparsification, intelligent sampling can also be applied temporally. 
Many \ac{CFD} datasets, whether statistically stationary, such as the OF2D, or transient, such as the SST-P1F4, consist of multiple snapshots of the solution as it evolves through time. These snapshots are extracted from the data producing simulations with a frequency determined \textit{a priori}, in some cases informed by knowledge of the physical time scales of the flow physics involved. However, this naive strategy often leads to training on time instances of the solution corresponding to similar points along the input \ac{PDF}, thus not contributing new information for surrogate model training. This phenomenon is especially true for solution trajectories which possess periodic characteristics, such as the vortex shedding behind a cylinder in the OF2D dataset. Intelligent subsampling can therefore be applied to the entire solution snapshot to discard ones that do not provide novel training data which helps better represent or expand the input \ac{PDF}. 

\renewcommand{\arraystretch}{1.5}
\begin{table*}[ht]
    \centering
    \caption{Neural network architectures used in the study. B is the batch size, T is input time sequence length, T' is predicted time horizon sequence length, C is the number of input variables, C' is the number of output variables, (H, W, D) is the (height, width, depth) of the grid, and N is the number of subsamples per timestep.}
    \begin{tabular}{lcc>{\raggedright\arraybackslash}p{5.5cm}p{3cm}}
        \hline
        \textbf{Architecture} & \textbf{Input Shape} & \textbf{Output Shape} & \textbf{Description} & \textbf{Input Data} \\ 
        \hline
        \textbf{LSTM} 
        & [B, T, C] 
        & [B, T', C'] 
        & Two LSTM layers, three dense layers & Subsampled Points \newline (Unstructured) \\
        \textbf{MLP-Transformer} 
        & [B, T, C, N] 
        & [B, T', C', H, W, D] 
        & MLP Encoder, Transformer Encoder, CNN Decoder (ConvTranspose3D) & Subsampled Points \newline (Unstructured) \\ % (8 heads, 6 layers)
        \textbf{CNN-Transformer} 
        & [B, T, C, H, W, D] 
        & [B, T', C', H, W, D] 
        & CNN Encoder (Conv3D), Transformer Encoder, CNN Decoder (Conv3D) & Extracted Hypercubes \newline (Structured) \\ % (8 heads, 6 layers)
        \hline
    \end{tabular}
    \label{tab:nn-architectures}
\end{table*}

\section{Model Training}
\label{sec:training}

Surrogate models replace traditional simulations with efficient, machine-learned approximations, either fully (e.g., FourCastNet~\cite{kurth2023fourcastnet}) or partially for components such as turbulence closures~\cite{beck2021perspective}. Unlike traditional surrogates, foundation models trained on extensive datasets offer generalization across multiple applications~\cite{bommasani2021opportunities}. However, their training is computationally demanding~\cite{cottier2024rising}. SICKLE addresses this challenge via intelligent subsampling for efficient model training, demonstrated across structured turbulence prediction, unstructured turbulence representation, and multiscale adaptive foundation models.

\subsection{Training Methodology}

The training component of SICKLE is built on the PyTorch deep learning framework~\cite{paszke2019pytorch}, in which we implement a modular, pluggable architecture for defining and testing neural network architectures. Users can specify architectures as a separate Python module, allowing for flexible experimentation with different configurations. 
In this paper, we explore three types of machine learning problems:

\begin{itemize}
    \item Baseline (\textit{full-full}) -- This is our baseline case with no sampling, where fully dense hypercubes are selected from the large simulation datasets.
    
    \item Sample reconstruction (\textit{sample-full}) -- This case involves learning a mapping from sparse, subsampled hypercubes to a fully dense target, e.g., sparse sensor reconstruction~\cite{malioutov2005sparse,manohar2018data}, such as extracting velocity or pressure profiles in the ocean using sparse flow probes. 
    
    \item Global prediction (\textit{sample-single}) -- In this case, we map many features to a single global prediction per timestamp, e.g., predicting drag on a cylinder given samples from the flowfield.
\end{itemize}

As summarized in Table~\ref{tab:nn-architectures}, we utilize three different types of neural network architectures in this study: (1) a simplified long short-term memory (LSTM) architecture that predicts a single scaler value over time (\textit{sample-single}), (2) a transformer-based multi-layer perceptron encoder that takes unstructured down-sampled data as input and predicts the full flowfield  (\textit{sample-full}), and (3) a transformer-based 3D convolutional neural network that takes structured hypercubes as inputs and also predicts the full flowfield (\textit{full-full}).

Scalable training is achieved via PyTorch’s Distributed Data Parallel module (\texttt{torch.distributed})~\cite{pytorch_ddp_class}, which enables multi-GPU and multi-node execution. For the purpose of enabling future studies on additional efficiency gains via reduced precision, SICKLE supports mixed-precision training via PyTorch Automatic Mixed Precision (\texttt{torch.amp}) by setting the \texttt{{-}-precision} flag to \texttt{fp16}, \texttt{bf16}, or \texttt{int8}. Moreover, SICKLE includes built-in support for scalable hyperparameter optimization using DeepHyper~\citep{balaprakash2018deephyper} via the \texttt{{-}-tune} option, which leverages Bayesian strategies to identify optimal neural network architectures and training configurations.

\subsection{Evaluation of Intelligent Subsampling on Trained Models}

We train the models in parallel across four AMD MI250X GPUs. Each GPU has two graphics compute dies (GCD), so we effectively use eight MPI processes per node. We train each model for 1000 epochs, with starting learning rate of 0.001, and use learning rate plateau with a patience of 20. We perform a 90:10 train:test split of the data before training. We use a batch size of 16. 

Initially, we investigated the maximum size of the hypercubes for training. We found that we could increase the size up to 32×32×32 for SST-P1, while keeping the training tractable. Because the time complexity of the attention mechanism in transformers is well known to be quadratic~\cite{ren2021combiner}, training becomes prohibitively slow when using larger than 32×32×32-sized hypercubes. Because of this issue, there has been considerable amount of research in developing more efficient implementations of transformers~\cite{tay2022efficient}, an interesting topic for future studies.

\subsubsection{Effect of sampling on model accuracy}

\begin{figure}
    \centering
    \includegraphics[width=\linewidth]{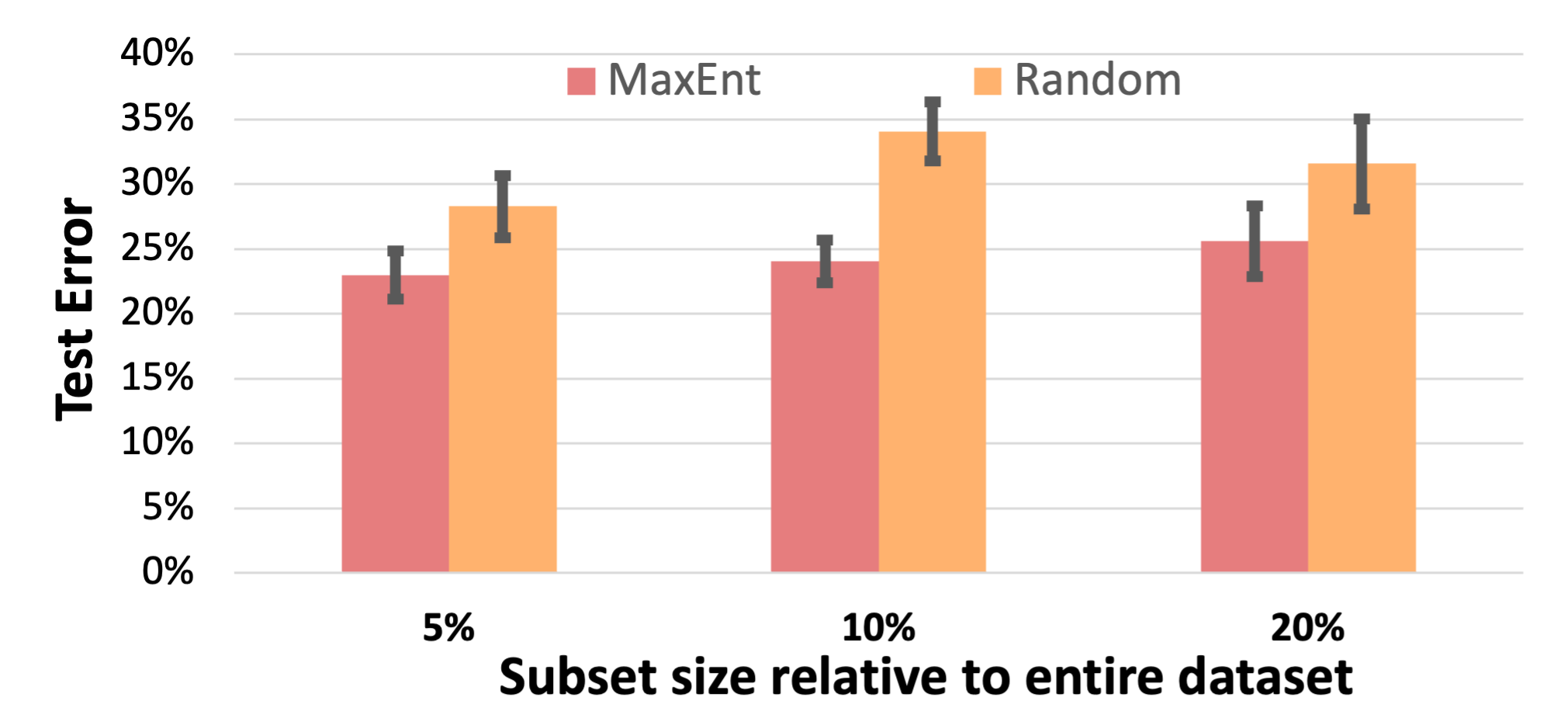}
    \caption{MaxEnt vs random sampling drag prediction surrogate model for OF2D case.}
    \label{fig:lstm}
\end{figure}

We compare the effect of different sampling strategies on model accuracy. Figure~\ref{fig:lstm} shows the test error of neural network surrogates trained to predict drag in a 2D flow over a cylinder, using either random or MaxEnt sampling. 
MaxEnt often produces more accurate and reproducible models than random sampling, particularly for specific target objectives such as drag, though random sampling performs competitively in many scenarios.

We also compare different combinations of hypercube extraction and point subsampling methods on on the SST-P1F4 and SST-P1F100 datasets which will be shown in Fig.~\ref{fig:energy}. The advantage of MaxEnt is most pronounced in datasets with high spatiotemporal variability, where uniform or random sampling is more likely to miss rare, information-rich regions, and will be further discussed in Sections \ref{sec:benchmarking} and \ref{sec:discussion}. 

\subsubsection{Towards a Foundation Model for Turbulence} % Pei

Developing a trustworthy foundation model for turbulence requires massive, diverse datasets from multiple sources. Prior work by Zhang et al.~\cite{zhang2024matey} has shown the benefits of training on heterogeneous datasets from different physical systems. In this work, we focus on using intelligent sampling to improve training efficiency on compute-intensive DNS data. The resulting improvements, as shown in Fig.~\ref{fig:matey} and discussed in Section~\ref{sec:benchmarking}, demonstrate that subsampling frameworks like SICKLE are highly effective in large-scale foundation model training. Other recent efforts, such as MATEY~\cite{zhang2024matey}, have begun to explore adaptive, multiscale modeling strategies for spatiotemporal physical systems. Our approach is complementary, focusing on data sparsification and sampling control as a scalable precursor to training. Future work could explore integrating intelligent sampling strategies like SICKLE into broader spatio-temporal foundation model frameworks that supports training across datasets of varying fidelity and scale.

\section{Efficiency and Scalability}
\label{sec:benchmarking}

We assess the efficiency and scalability of our subsampling methods using Frontier's Cray Power Management counters, which offer reliable energy measurements. We approximate the total training cost by:

\begin{equation}
    \text{Cost to Train} \approx \mathcal{O}(c(m)) + \mathcal{O}(m \cdot p \cdot e)
    \label{eq:training-cost}
\end{equation}

\noindent where \(c(m)\) is the initial sampling cost, $m$ the number of training samples, $p$ model parameters, and $e$ epochs. Efficiency gains from subsampling thus reduce both per-epoch computational costs and amortize the initial sampling overhead.

\begin{figure}
    \centering
    \includegraphics[width=0.95\linewidth]{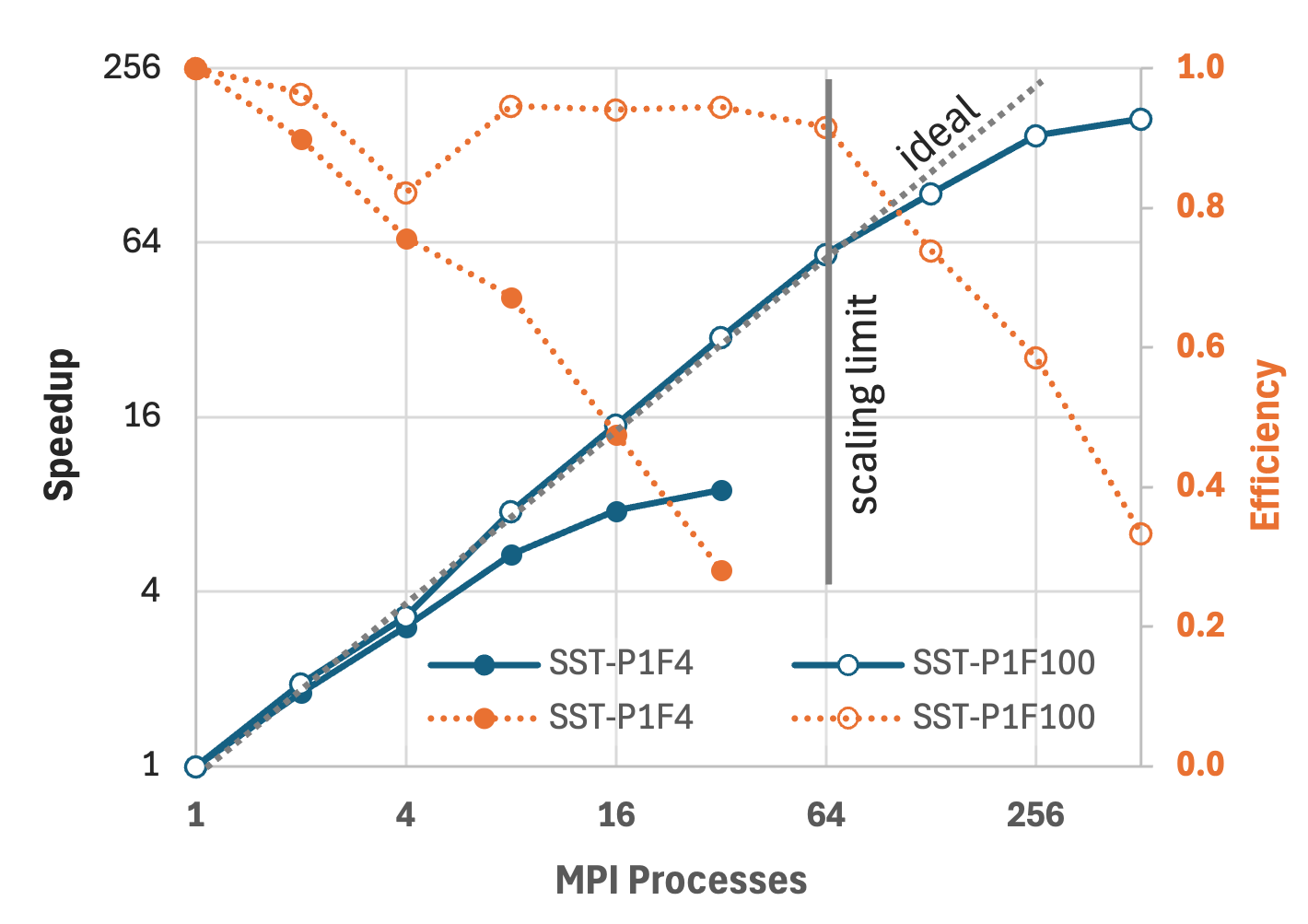}
    \caption{MaxEnt parallel scalability of SST-P1F4 and SST-P1F100 cases. The vertical line marks the scaling limit (knee point), beyond which efficiency drops sharply as the dataset becomes too thinly distributed across MPI processes to keep them fully utilized.}
    \label{fig:scalability}
\end{figure}

Figure~\ref{fig:scalability} demonstrates \textit{scalability} of MaxEnt sampling up to 512 MPI processes. SST-P1F100 scales effectively to 64 MPI processes before efficiency declines, achieving 171× speedup at 512 MPI processes. SST-P1F4 shows limited scaling due to its smaller size. 

\begin{figure}
    \centering
    \includegraphics[width=\linewidth]{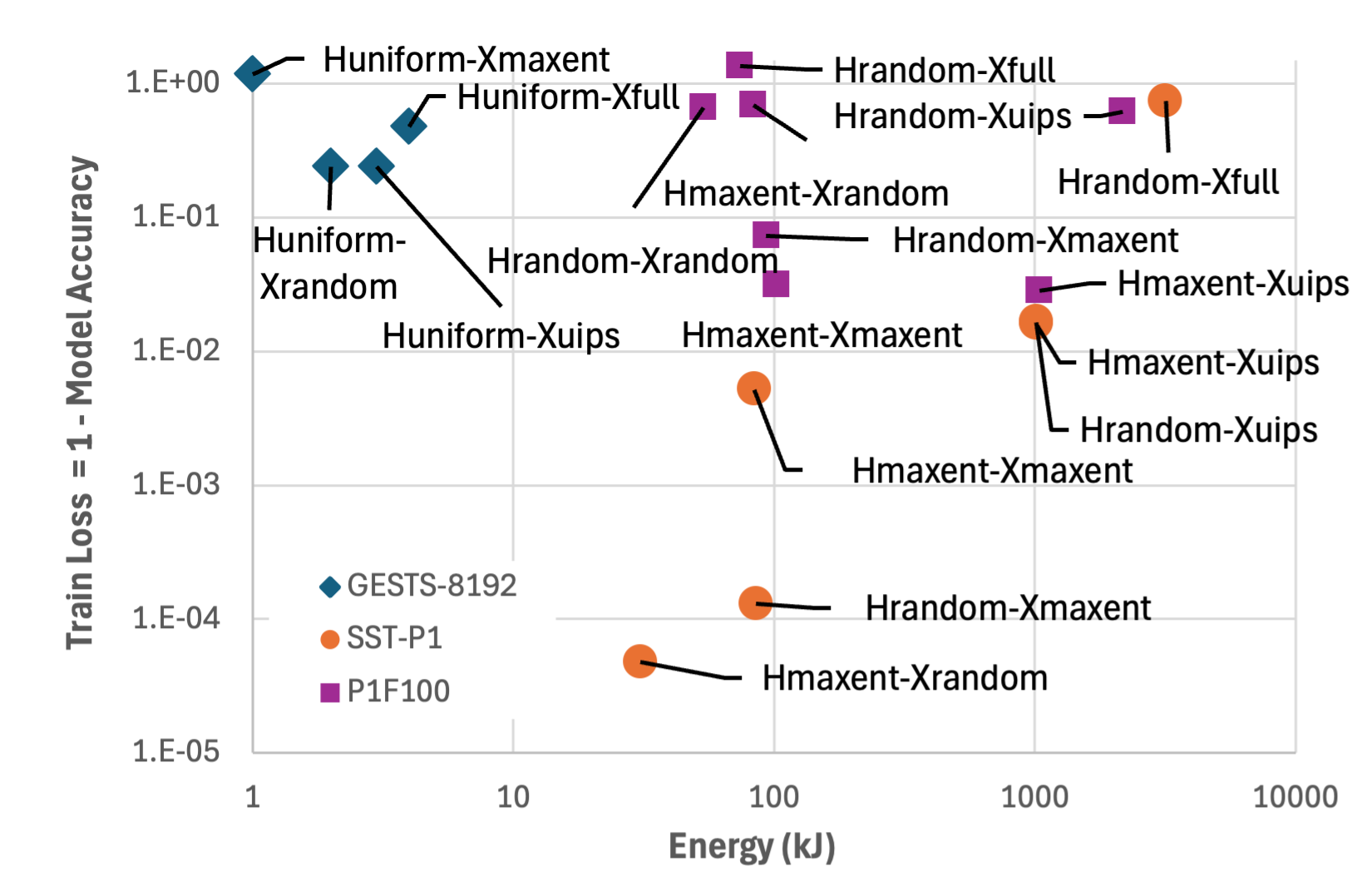}
    \caption{Training loss vs energy cost (lower-left is optimal) for different sampling methods.}
    \label{fig:energy}
\end{figure}

Figure~\ref{fig:energy} compares \textit{accuracy} and \textit{energy} use for different sampling methods on the GESTS-8192, SST-P1, and SST-P1F100 datasets. MaxEnt generally provides lower training loss at reduced energy cost relative to UIPS and full sampling, particularly for the stratified turbulence cases (SST-P1). For example, in one SST-P1 case MaxEnt required about 85 kJ, compared to 1,000 kJ for UIPS and 3,183 kJ for full sampling—38× more energy than MaxEnt. However, the advantage is less pronounced for isotropic turbulence (GESTS-8192), where all methods yield relatively high loss despite low energy use. These results suggest that the benefits of intelligent subsampling are most evident for large, anisotropic datasets with significant redundancy, while gains are more modest for isotropic flows. Overall, energy savings from subsampling increase with dataset size, 
but the optimal method depends on the interplay between dataset characteristics, model architecture, and reproducibility requirements.

\begin{figure}
    \centering
    \includegraphics[width=\linewidth]{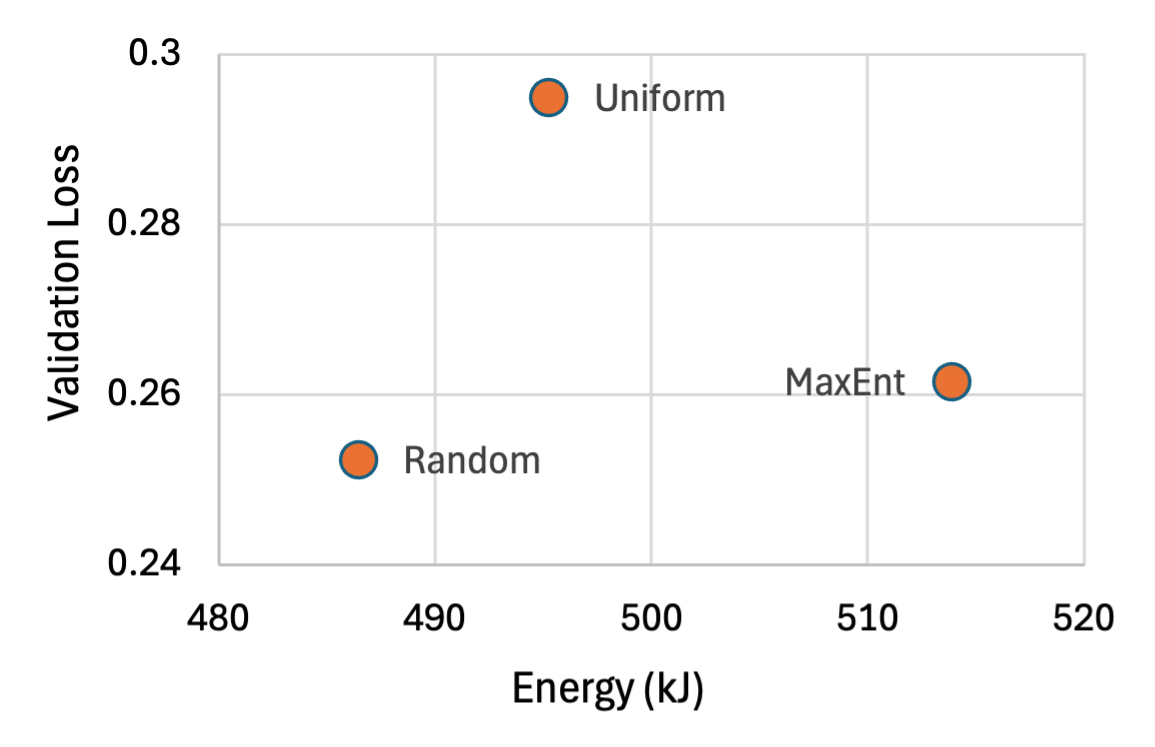}
    \caption{Model accuracy and energy consumption of the MATEY foundation model trained with the SST-P1F4 data at $10\%$ sampling rate.}
    \label{fig:matey}
\end{figure}

As a preliminary evaluation of the SICKLE framework for improving foundation model training, we apply it on the MATEY model trained on the SST-P1F4 data for 50 epochs using a sampling rate of $10\%$. Figure~\ref{fig:matey} reports the validation loss and total energy consumed with three sampling strategies: uniform, random, and MaxEnt. 
In this case random sampling attained the lowest validation loss (0.252)
and also used the least energy (486 kJ), MaxEnt consumed 514kJ at a validation loss of 0.262, while uniform sampling consumed 495 kJ at a considerably higher validation loss of 0.295. This was an initial study, which requires a more systematic and thorough analysis of training at different sampling rates and with different datasets to better quantify the trade-offs between accuracy and energy use for foundation model training. 

\section{Discussion}
    \label{sec:discussion}

    We initially hypothesized that intelligent subsampling techniques, such as MaxEnt, would generally outperform naive methods like random sampling. Surprisingly, we found that random sampling performs quite well in many scenarios, which is consistent with prior work in \acp{LLM}, which noted that ``nearly all methods struggled to significantly outperform random selection when dealing with such large-scale data pools'' \cite{xia2024rethinking}. 
    On the other hand, Hochlehnert et al.~\cite{hochlehnert2025sober} note that \acp{LLM} are ``highly sensitive to subtle implementation choices,'' citing sampling variance (primarily due to randomness) as a root cause of irreproducibility. MaxEnt exhibits less variance and is therefore more reproducible than random sampling (see Fig.~\ref{fig:lstm}).

    As opposed to text corpora, scientific datasets such as turbulence are highly nonlinear, requiring more informed sampling. Our results indicate that MaxEnt offers notable advantages for anisotropic flows—those with strong directional gradients—as evidenced by its performance on the SST and OF2D datasets (see Fig.~\ref{fig:pdf}). In these cases, MaxEnt effectively captures the essential structure of the flowfield with significantly fewer samples, resulting in accurate and efficient models. For isotropic turbulence, such as the GESTS datasets, performance differences between sampling strategies were less pronounced. Future work may explore whether alternative variable selections or wavenumber-based analyses further clarify these outcomes. Ultimately, deploying MaxEnt in production settings will require a systematic examination of variable choice to fully leverage its potential across different turbulence regimes.

    Of course, in the limit of large sample sizes, random sampling is expected to converge to the true \ac{PDF}, eventually matching or even exceeding MaxEnt’s performance. Random sampling converges at a rate of \( \mathcal{O}(n^{-4/5}) \) because estimating a full distribution nonparametrically (e.g., via KDE) is harder than estimating just a few parameters—it requires more data to achieve the same level of accuracy \cite{estimatingDistributions2009}. Yet this is precisely where MaxEnt's value lies: in its ability to provide near-optimal coverage with significantly fewer points in settings where structure (e.g., anisotropy) can be exploited. This efficiency comes at a computational cost of performing a cluster analysis, but when sample budgets are tight or energy constraints are a priority, the trade-off may be well worth it (see Fig.~\ref{fig:energy}).

\section{Conclusions and Future Work}
\label{sec:conclusions}

We set out with the hypothesis that maximum entropy–based subsampling could yield more accurate models with greater computational and energy efficiency than training on full or randomly selected datasets. Our findings show that this hypothesis is only \textit{partially} supported. This nuance points to the need for systematic evaluation of when intelligent sampling is most beneficial. Even so, our results confirm that intelligently curated sparse datasets can achieve accuracy comparable to, and in some cases better than, models trained on full datasets, while dramatically reducing training energy—up to two orders of magnitude in some cases.
We introduced SICKLE, an open-source framework utilizing maximum entropy-based sampling. SICKLE first extracts optimal hypercubes, then selects high-entropy points within each cube, proving effective for surrogate and scientific foundation models.

Our evaluations using direct numerical simulation datasets, ranging from megabytes to hundreds of terabytes, showed that while random sampling performed well in many cases, SICKLE with MaxEnt sampling occasionally outperformed both random and phase-space sampling methods, particularly in anisotropic turbulence cases. In these settings—where directional gradients and redundancy made intelligent subsampling especially impactful—MaxEnt provided more reproducible improvements, delivering higher accuracy at lower energy cost (in some instances consuming less than 1\% of the baseline). By leveraging underlying data structure, MaxEnt can achieve broader coverage with fewer samples. Although clustering incurs additional computational overhead, this trade-off can be worthwhile in data- or energy-constrained settings (Fig.~\ref{fig:energy}).

Current limitations include non-optimized raw data ingestion and opportunities to improve parallel scalability across spatial and temporal dimensions. Looking forward, we envision SICKLE integrated into AI-coupled HPC workflows, with several promising extensions:

\begin{itemize}
\item Adaptive temporal sampling responsive to transient phenomena and evolving model uncertainty.
\item Integration with in-situ, streaming, and online training frameworks like SmartSim~\cite{Partee_2022}.
\item Support for federated learning across distributed HPC facilities using frameworks like APPFL~\cite{ryu2022appfl}.
\item Enhanced visualization and analysis tools compatible with VTK and ParaView.
\item Applications across other scientific domains, notably climate and fusion.
\item Integration into spatio-temporal foundation model frameworks such as MATEY~\cite{zhang2024matey}. 
\end{itemize}

We hope this work encourages broader investigation of intelligent subsampling as a promising approach for efficient and scalable training of scientific foundation models.

\begin{acks}
This research was sponsored by and used resources of the Oak Ridge Leadership Computing Facility (OLCF), a DOE Office of Science User Facility at Oak Ridge National Laboratory (ORNL), supported by the U.S. Department of Energy under Contract No. DE-AC05-00OR22725. 
In addition, it was supported by the Artificial Intelligence Initiative as part of the Laboratory Directed Research and Development (LDRD) Program at ORNL. 
Portions of this work also used resources of the Argonne Leadership Computing Facility (ALCF), a DOE Office of Science User Facility at Argonne National Laboratory, and were supported by the DOE Office of Science, Advanced Scientific Computing Research Program, under Contract No. DE-AC02-06CH11357.
We would also like to thank a number of colleagues for helpful information and advice along the way, including: Malik Hassanaly for help with UIPS, as well as Shantenu Jha, Joel Bretheim, Andrew Shao,   Trey White, and Bronson Messer for providing feedback on the manuscript.
The authors acknowledge the use of ChatGPT (version 4.5, OpenAI) for language editing, LaTeX formatting assistance, and structural suggestions.
\end{acks}

\balance

\bibliographystyle{ACM-Reference-Format}
\bibliography{refs}

\clearpage
\appendix
\section{Appendix: Artifact Description}

\subsection{Paper's Main Contributions}

% \subsection*{Expected Results}l{
% Provide a list of all main contributions of the paper.
% }

We provide the following contributions in our paper:

\begin{description}
\item[$C_1$] Development of the SICKLE framework for intelligent data subsampling, including support for random, maximum entropy (MaxEnt) and phase-space sampling (UIPS) methods, among others. 
\item[$C_2$] Demonstration that intelligent subsampling improves model accuracy and reproducibility across multiple datasets.
\item[$C_3$] Benchmarking of energy efficiency versus accuracy sampling performance of different sampling on large-scale scientific workloads.
\item[$C_4$] Benchmarking parallel scalability of SICKLE.  
\end{description}

\subsection{Computational Artifacts}

% \subsection*{Expected Results}l{
% List the computational artifacts related to this paper along with their respective DOIs. Note that all computational artifacts may be archived under a single DOI.
% }

We have combined all the computational artifacts under a single DOI listed here:

\begin{description}
\item[$A_1$] SICKLE framework and all associated datasets used in this paper are available at: \url{https://doi.org/10.5281/zenodo.15226079}
\end{description}

% \subsection*{Expected Results}l{
% Provide a table with the relevant computational artifacts,
% highlight their relation to the contributions (from above) and
% point to the elements in the paper that are reproducible by each artifact, e.g.,
% which figures or tables were generated with the artifact.
% }

\noindent The following table relates the computational artifacts to their
contributions and related paper elements: \vspace{0.1in}

\begin{tabular}{rll}
\toprule
Artifact ID & Contributions & Description \\
\midrule
$A_1$ & $C_1$ & Figures 3, 4, 5 \\
      & $C_2$ & Figure 6 \\
      & $C_3$ & Figure 7 \\
      & $C_4$ & Figure 8 \\
\bottomrule
\end{tabular}

%%%%%%%%%%%%%%%%%%%%%%%%%%%%%%%%%%%%%%%%%%%%%%%%%%%%%%%%
\section{Artifact Identification}
%%%%%%%%%%%%%%%%%%%%%%%%%%%%%%%%%%%%%%%%%%%%%%%%%%%%%%%%

% \subsection*{Expected Results}l{
% Provide the following six subsections for each computational artifact $A_i$.
% }

Here, we provide $A_1$ artifact's relation to contributions, expected results, expected reproduction time, artifact setup, artifact execution, and artifact analysis. 

\subsection{Computational Artifact}

\subsection*{Relation To Contributions}

% \subsection*{Expected Results}l{
%     Briefly explain the relationship between the artifact and contributions.
% }

Here, we briefly explain the relationship between artifact $A_1$ and the contributions. The artifact provides the SICKLE framework and datasets required to reproduce the experimental results supporting all three contributions. For $C_1$, the artifact enables implementation and comparison of sampling methods (random, UIPS, MaxEnt). For $C_2$, it includes the training pipeline and configuration files used to evaluate model accuracy and reproducibility across datasets. For $C_3$, the artifact supports energy and scalability benchmarking using subsampled datasets and multi-process configurations.

\subsection*{Expected Results}

% \subsection*{Expected Results}l{
% Provide a higher level description of what outcome to expect from the corresponding experiments. Provide an explanation of how the results substantiate the main contributions.
% }

% Here, we provide a higher level description of what outcome to expect from the corresponding experiments, and explain how the results substantiate the main contributions. 

This section outlines a high-level summary of the experimental results and their alignment with the main contributions.

$C_1$: The experiments show that MaxEnt sampling outperforms random and UIPS methods in capturing complex flow structures, particularly in the tails of the data distribution. Fig. 3 visualizes wake-region fidelity in OF2D; Fig. 4 illustrates UIPS sensitivity to flowfield complexity in TC2D vs. SST-P1F4; and Fig. 5 demonstrates PDF alignment across three datasets, where MaxEnt achieves the best match, especially in the tails.

$C_2$: Figure 6 demonstrates that models trained on MaxEnt-sampled datasets yield 5–10\% lower prediction error and improved reproducibility compared to those trained on random samples, validating the benefit of intelligent subsampling for surrogate modeling.

$C_3$: Figures 7 and 8 show that MaxEnt enables models to achieve lower training loss while using significantly less energy compared to both UIPS and full datasets. SICKLE’s parallel implementation demonstrates strong scaling behavior for large datasets like SST-P1F100, with performance tapering for smaller datasets due to communication overheads.

% \artsampl{
% Algorithm A should be faster than Algorithms C and B in all GPU scenarios.
% }

\subsection*{Expected Reproduction Time (in Minutes)}

% \subsection*{Expected Results}l{
% Estimate the time required to reproduce the artifact, providing separate estimates for the individual steps: Artifact Setup, Artifact Execution, and Artifact Analysis.
% }

% \artsampl{
% The expected computational time of this artifact on GPU X is 20~min.
% }

\textbf{Artifact Setup:}  
Initial setup includes cloning the SICKLE repository, installing dependencies via Conda or Spack, and downloading datasets from the Zenodo archive. On Frontier, this setup process is expected to take approximately 30 minutes. All experiments are designed to run on a single node (4x AMD MI250X GPUs, 64-core AMD EPYC 7713 “Trento” CPU).

\noindent \textbf{Artifact Execution and Analysis (per Figure):}

\begin{itemize}
    \item \textbf{Figure 3} (OF2D sampling comparison):\\
    Execution: $\sim$2 minutes (CPU-only); Analysis: $\sim$10m (Excel)

    \item \textbf{Figure 4} (UIPS sampling on TC2D and SST-P1F4):\\
    Execution: $\sim$5 minutes; Analysis: $\sim$5 minutes (Matplotlib)

    \item \textbf{Figure 5} (PDF comparison across OF2D, SST, GESTS):\\
    Execution: $\sim$8 minutes (1 GPU); Analysis: $\sim$5 minutes (Matplotlib)

    \item \textbf{Figure 6} (OF2D surrogate model accuracy):\\
    Execution: $\sim$10 minutes (1 GPU); Analysis: $\sim$15-20 minutes (Excel)

    \item \textbf{Figure 7} (SST training loss vs. energy use):\\
    Execution: $\sim$12 minutes (2 GPUs); Analysis: $\sim$15-20 minutes (Excel)

    \item \textbf{Figure 8} (Parallel scalability on SST-P1F100):\\
    Execution: $\sim$26 minutes (using up to 512 MPI ranks); Analysis: $\sim$15-20 minutes (Excel)
\end{itemize}

\vspace{0.5em}
Total estimated time to reproduce all figures: \textbf{1.5–2 hours}, assuming access to a single Frontier node.

\subsection*{Artifact Setup - Hardware}

% \subsection*{Expected Results}l{
% Specify the hardware requirements and dependencies (e.g., a specific interconnect or GPU type is required).
% }

The artifact is designed to run on a single node of the Frontier supercomputer, which includes:
\begin{itemize}
    \item 1x AMD EPYC 7713 “Trento” CPU (64 cores, 2.0 GHz)
    \item 4x AMD Instinct MI250X GPUs
    \item High-speed HPE Slingshot interconnect (used for scalability tests in Fig.~8)
\end{itemize}
Most figures can be reproduced using a single node (Figs. 3-7). Scalability experiments (Fig. 8) require 32 MPI ranks across per node up to 16 nodes.

\subsection*{Artifact Setup (incl. Inputs)}part{Software}

% \subsection*{Expected Results}l{
% Introduce all required software packages, including the computational artifact. For each software package, specify the version and provide the URL.
% }

The SICKLE framework is available at: \url{https://code.ornl.gov/autosm/sickle}. SICKLE is implemented in Python and requires the following Python packages: 

\vspace{0.5em}
\noindent
% \textbf{Core dependencies:}
\begin{itemize}
  \item \texttt{numpy} 1.23.5 – \url{https://numpy.org}
  \item \texttt{scipy} 1.14.0 – \url{https://scipy.org}
  \item \texttt{pandas} 2.2.2 – \url{https://pandas.pydata.org}
  \item \texttt{matplotlib} 3.7.1 – \url{https://matplotlib.org}
  \item \texttt{scikit-learn} 1.2.2 – \url{https://scikit-learn.org}
  \item \texttt{pyyaml} 6.0.2 – \url{https://pyyaml.org}
  \item \texttt{fluidfoam} 0.2.5 – \url{https://github.com/fluiddyn/fluidfoam}
  \item \texttt{cftime} 1.6.4 – \url{https://unidata.github.io/cftime}
  \item \texttt{netCDF4} 1.7.1.post1 – \url{https://github.com/Unidata/netcdf4-python}
  \item \texttt{h5py} 3.11.0 – \url{https://www.h5py.org}
  \item \texttt{mpi4py} 3.1.4 – \url{https://mpi4py.readthedocs.io}
  \item \texttt{pydmd} 0.4.1 – \url{https://mathlab.github.io/PyDMD}
  \item \texttt{absl-py} 2.1.0 – \url{https://github.com/abseil/abseil-py}
  \item \texttt{sympy} 1.13.1 – \url{https://www.sympy.org}
  \item \texttt{networkx} 3.3 – \url{https://networkx.org}
  \item \texttt{requests} 2.32.3 – \url{https://docs.python-requests.org}
\end{itemize}

These packages are compatible with Python 3.10+ and have been tested on Frontier using ROCm-compatible builds, and have also been tested to work on macOS Sequioa (using CPU version of PyTorch).
All packages were installed using \texttt{pip} and are compatible with ROCm-based systems. Users may optionally adapt the list to create a Conda environment for ease of deployment. 
A requirements file are included to simplify environment creation.

\subsection*{Artifact Setup (incl. Inputs)}part{Datasets / Inputs}

% \subsection*{Expected Results}l{
% Describe the datasets required by the artifact. Indicate whether the datasets can be generated, including instructions, or if they are available for download, providing the corresponding URL.
% }

Links to all datasets used in this paper are archived at: \url{https://doi.org/10.5281/zenodo.15226079}. 
The archive includes preprocessed subsets of the following datasets:
\begin{itemize}
    \item TC2D (2D turbulent combustion)
    \item OF2D (2D flow over cylinder, OpenFOAM-generated)
    \item SST-P1F4 / SST-P1F100 (forced stratified turbulence)
    \item GESTS-2048 / GESTS-8192 (forced isotropic turbulence)
\end{itemize}
Each dataset has a custom dataloader provided to read the dataset, under the `dataloaders' directory.
% All inputs are provided in `.npy` or `.npz` format and include instructions for how they are used in training or sampling.

\subsection*{Artifact Setup (incl. Inputs)}part{Installation and Deployment}

% \subsection*{Expected Results}l{
% Detail the requirements for compiling, deploying, and executing the experiments, including necessary compilers and their versions.
% }

No compilation is required beyond setting up the Python environment. Users are expected to load ROCm-compatible modules on Frontier and install Python dependencies via Conda or Spack. A setup script (`contrib/env-frontier') is provided to initialize the environment, which contains the following: 

\begin{verbatim}
# Setup Python environment
source /path/to/sickle/venv/pyt/bin/activate
# Set environment variables for PyTorch
WORLD_SIZE=$((SLURM_NTASKS))
NODE_RANK=$SLURM_NODEID
export MASTER_ADDR=$(hostname -i)
export NCCL_SOCKET_IFNAME=hsn0
export MASTER_PORT=3442
export PYTORCH_ROCM_ARCH=gfx90a
# Needed to bypass MIOpen, Disk I/O Errors
export MIOPEN_USER_DB_PATH="/tmp/my-miopen-cache"
export MIOPEN_CUSTOM_CACHE_DIR=${MIOPEN_USER_DB_PATH}
rm -rf ${MIOPEN_USER_DB_PATH}
mkdir -p ${MIOPEN_USER_DB_PATH}
# Load required modules
module load rocm/6.3.1 libfabric/1.22.0
\end{verbatim}

\noindent MPI-based scalability experiments rely on `torch.distributed'. See \url{https://pytorch.org/docs/stable/distributed.html} for more info. 

% \subsection*{Artifact Setup (incl. Inputs)}part{Hardware}

% \subsection*{Expected Results}l{
% Specify the hardware requirements and dependencies (e.g., a specific interconnect or GPU type is required).
% }

% \subsection*{Artifact Setup (incl. Inputs)}part{Software}

% \subsection*{Expected Results}l{
% Introduce all required software packages, including the computational artifact. For each software package, specify the version and provide the URL.
% }

% \subsection*{Artifact Setup (incl. Inputs)}part{Datasets / Inputs}

% \subsection*{Expected Results}l{
% Describe the datasets required by the artifact. Indicate whether the datasets can be generated, including instructions, or if they are available for download, providing the corresponding URL.
% }

% \subsection*{Artifact Setup (incl. Inputs)}part{Installation and Deployment}

% \subsection*{Expected Results}l{
% Detail the requirements for compiling, deploying, and executing the experiments, including necessary compilers and their versions.
% }

\subsection*{Artifact Execution}

% \subsection*{Expected Results}l{
% Provide an abstract description of the experiment workflow of the artifact. It is important to identify the main tasks (processes) and how they depend on each other.

% A workflow may consist of three tasks: $T_1, T_2$, and $T_3$. The task $T_1$ may generate a specific dataset. This dataset is then used as input by a computational task $T_2$, and the output of $T_2$ is processed by another task $T_3$, which produces the final results (e.g., plots, tables, etc.). State the individual tasks $T_i$ and provide their dependencies, e.g., $T_1 \rightarrow T_2 \rightarrow T_3$.

% Provide details on the experimental parameters. How and why were parameters set to a specific value (if relevant for the reproduction of an artifact), e.g., size of dataset, number of data points, input sizes, etc. Additionally, include details on statistical parameters, like the number of repetitions.
% }

\textbf{Experiment Workflow:}

The experimental workflow consists of the following main tasks:

\begin{itemize}
    \item[$T_1$] \textbf{Sampling:} Generate training subsets using random, UIPS, and MaxEnt sampling strategies via the SICKLE framework. Sampling was performed using the following command: \\
    \texttt{srun -n 32 python -u subsample.py case.yaml}. 
    \item[$T_2$] \textbf{Model Training:} Train neural networks (e.g., LSTM, CNN-Transformer, MLP-Transformer) on each sampled subset using specified loss functions and hyperparameters. Training was perfromed using the following command: \\
    \texttt{srun -n 8 python -u train.py case.yaml}
    \item[$T_3$] \textbf{Evaluation and Analysis:} Compute predictive accuracy (1 - loss), generate PDFs, aggregate energy usage, and produce final plots. Energy is listed under ``Total Energy Consumed'' in the output of $T_1$ and $T_2$, and ``Elapsed Time'' is also measured. The loss value on the test set is labeled as ``Evaluation on test set''. 
\end{itemize}

The dependency structure is linear: $T_1 \rightarrow T_2 \rightarrow T_3$. All tasks are driven by YAML configuration files included in the artifact.

\textbf{Experimental Parameters:}

\begin{itemize}
\item Sampling sizes were set to fixed fractions of the full dataset to compare performance across scales. Unless otherwise stated we used 10\% of the total points, which ends up being 3277 points per hypercube for SST and GESTS datasets, and 1080 for the OF2D dataset. 
\item Models were trained for a fixed number of epochs (e.g., 1000) using Adam optimizer with learning rate plateau using patience of 20. 
\item Each experiment was repeated 3 times using different random seeds to compute mean accuracy and standard deviation.
\item Parallel scalability tests varied the number of MPI ranks from 1 to 512 doubling by factors of 2 to evaluate performance on large datasets (SST-P1F100).
\item PDF comparisons were binned using a fixed bin size of 100 across all datasets for consistency.
\end{itemize}

All experimental parameters and random seeds are defined in the config files accompanying each figure script. A sample YAML file for SST-P1F4 is provided here as an example:

\begin{verbatim}
shared:
  dims: 3
  dtype: sst-binary
  input_vars: [u, v, w, r]
  output_vars: p
  cluster_var: pv
  nx: 514
  ny: 512
  nz: 256
  gravity: z
   fileprefix: "SST-P1-H{hypercubes}-C{num_hypercubes}"+\
              "-X{method}-ns{num_samples}-window{window}"
subsample:
  hypercubes: maxent
  num_hypercubes: 32
  method: maxent
  path: /path/to/P1F4R32_testing/raw_data/
  num_samples: 3277
  num_clusters: 20
  nxsl: 32
  nysl: 32
  nzsl: 32
train:
  epochs: 1000
  batch: 16
  target: p_full
  window: 1
  arch: MLP_transformer
  sequence: true
\end{verbatim}

A sample Slurm script for performing multiple sampling and training runs, such as was used for Figures 7 \& 8 is provided here:

\begin{verbatim}
# Setup environment
SRC="/path/to/sickle"
. $SRC/contrib/environment

# Define the list of cases
CASES=("Hmaxent-Xmaxent-32" \
       "Hmaxent-Xuips-32" \
       "Hrandom-Xfull" \
       "Hrandom-Xmaxent-32" \
       "Hrandom-Xuips-32")

# Define the run directory and source path
RUNDIR="$MEMBERWORK/xyz123/sickle/${SLURM_JOB_ID}"
mkdir -p $RUNDIR "$RUNDIR/snapshots" "$RUNDIR/plots"

# Copy all case files to the run directory
for CASE in "${CASES[@]}"; do
    echo "Copying case file: $CASE.yaml"
    cp $SRC/contrib/configs/SST/P1/$CASE.yaml $RUNDIR
done

# Copy the slurm.sh script for reproducibility
echo "Copying slurm.sh to $RUNDIR"
cp "${BASH_SOURCE[0]}" "$RUNDIR"

# Change directory to the run directory once
cd "$RUNDIR" || exit

# Initialize the counter variable
count=0

# Loop over each case and execute the commands
for CASE in "${CASES[@]}"; do
    echo "Processing case: $CASE"

    ### SUBSAMPLING
    time srun -N "$SLURM_NNODES" -n 32 python -u \ 
              "$SRC/subsample.py" "$CASE.yaml" \
              --output_dir "$RUNDIR/snapshots" >& 
              "$RUNDIR/subsample${count}.out"

    ### TRAINING
    time srun -N "$SLURM_NNODES" --ntasks-per-node=8 \
              python -u "$SRC/train.py" --plot \
              --output_dir "$RUNDIR/snapshots" \
              "$CASE.yaml" >& "$RUNDIR/train${count}.out"

    echo "Finished processing case: $CASE"

    # Increment the counter
    count=$((count + 1))
done
\end{verbatim}

\subsection*{Artifact Analysis (incl. Outputs)}

\textbf{Artifact Analysis and Expected Results:}

Below we provide figure-specific instructions for reproducing each result, expected outcomes, and explain their correlation to the paper's contributions.

\vspace{0.5em}

\noindent \textbf{Figure 3: Sampling visualizations (OF2D)}

\noindent \textit{Reproduction Instructions:}
\begin{verbatim}
$ tar xvfz data.tgz
$ python subsample.py --dtype openfoam --path ./data \
  -cv wz --method [full|random|uips|maxent] --plot \
  --timesteps 97
\end{verbatim}
Run the command four times, changing \texttt{{-}-method} each time to be \texttt{full}, \texttt{random}, \texttt{uips}, and \texttt{maxent}, respectively. 

\noindent \textit{Expected Output:}

Generates \texttt{subsample\_plot\_t0097.png} for each method, showing stacked sampling visualizations. MaxEnt should best capture wake structures.

\noindent \textit{Correlation to Contributions:} 

Supports $C_1$ by visually demonstrating sampling effectiveness for different sampling methods. 

\vspace{0.5em}

\noindent \textbf{Figure 4 (Left): UIPS (TC2D)}

\noindent \textit{Reproduction Instructions:}
\begin{verbatim}
$ git clone https://github.com/NREL/phase-space-sampling
$ cd phase-space-sampling
$ python setup.py install
$ python tests/main_from_input.py -i uips/inputs/input2D
$ python postProcess/visualizeDownSampled_subplots.py \
    -i uips/inputs/input2D
\end{verbatim}

\noindent \textit{Expected Output:}

Figure located at \path{Figures/downSampledData_10000_it1.png}, showing UIPS gives good, uniform sampling performance for TC2D.

\noindent \textit{Correlation to Contributions:} 

Demonstrates sampling variability supporting $C_1$.

\vspace{0.5em}

\noindent \textbf{Figure 4 (Right): UIPS (SST-P1F4)}

\noindent \textit{Reproduction Instructions:}
From within the \path{sickle/uips} directory, run:
 
\begin{verbatim}
$ python -u sst_to_npy.py --dims 3 --dtype sst-binary \
  --path ~/data/P1F4R32_nx512ny512nz256_6vars/ \
  --num_hypercubes 1 --noseed --plot \
  --input_vars u v w r --output_vars p --cluster_var pv \
  --nx 514 --ny 512 --nz 256 --gravity z \
  --nxsl 128 --nysl 128 --nzsl 64
\end{verbatim}

\noindent Now we can use the phase-space-sampling code to analyze 
a single timestep of our SST-P1F4 dataset, as follows:

\begin{verbatim}
$ python phase-space-sampling/tests/main_from_input.py \
  -i input4D
$ python phase-space-sampling/postProcess/ \
  visualizeDownSampled_subplots.py -i input4D
\end{verbatim}

\noindent \textit{Expected Output:}

Figure \texttt{downSampledData\_10000\_it1.png}, showing UIPS does not do as well on 3D complex flowfields, such as the SST-P1F4 3D forced stratified turbulence case. The sampled points do not provide uniform coverage of the feature space. 

\noindent \textit{Correlation to Contributions:} 

Reinforces $C_1$ by highlighting UIPS limitations, and why we should explore another method such as MaxEnt. 

\vspace{0.5em}

\noindent \textbf{Figure 5: PDF comparisons}

\noindent \textit{Reproduction Instructions:}
Before running each command, set \texttt{ts} and \texttt{ymax} 
accordingly within \path{compare_methods.py}.

\noindent For Fig. 5a (OF2D), set \texttt{ts=97} and \texttt{ymax=1000}, then run:

\begin{verbatim}
$ python compare_methods.py --dtype openfoam \
  --path ~/data/cylinder -cv wz -ns 1080
\end{verbatim}

\noindent For Fig. 5b (SST-P1F4), set \texttt{ts=0} and \texttt{ymax=10}, then run:

\begin{verbatim}
$ python compare_methods.py --dtype sst-binary \
  --path ~/data/P1F4R32_nx512ny512nz256_6vars/ \
  --num_hypercubes 1 -nc 20 -ns 3277 --input_vars u v w r \
  --output_vars p --cluster_var pv --timesteps 28.44 \
  --nx 514 --ny 512 --nz 256 --gravity z --nxsl 32 \
  --nysl 32 --nzsl 32
\end{verbatim}

\noindent For Fig. 5c (GESTS-2048), set \texttt{ts=0} and \texttt{ymax=100}, then run:

\begin{verbatim}
$ python compare_methods.py --dtype gests \
  --path ~/data/GESTS -nc 20 -ns 3277 \
  --input_vars velocity dissipation \
  --output_vars pressure --cluster_var enstrophy \
  --nx 1024 --ny 1024 --nz 1024
  --nxsl 32 --nysl 32 --nzsl 32
\end{verbatim}

\noindent \textit{Expected Output:}

PDF plots showing how MaxEnt outperforms other methods in tail representation are saved in \path{plots/subsampling_methods_histograms*.png}.

\noindent \textit{Correlation to Contributions:} 

Confirms $C_1$ by quantifying sampling effectiveness.

\vspace{0.5em}

\noindent \textbf{Figure 6: Surrogate model accuracy study (OF2D)}

\noindent \textit{Reproduction Instructions:}
Install Docker from \url{docker.com}. Follow detailed steps to convert the grid from body-fitted grid to cartesian grid:
\begin{verbatim}
$ docker pull openfoam/openfoam7-paraview56
$ docker run -it -v $HOME/data:/home/openfoam \
                    openfoam/openfoam7-paraview56
>> foamToVTK 
\end{verbatim}
Here we assume that the previously downloaded \texttt{data.tgz} is located in the home directory. 
Next, install ParaView from \url{https://www.paraview.org/download/} and run the following:
\begin{verbatim}
$ pvpython interpolate.py
\end{verbatim}
Note, on macOS systems \texttt{pvpython} is located in \path{/Applications/ParaView-5.X.0.app/Contents/bin}, 
where X is the version number, e.g., 13. 

Run the following subsampling commands for three replicates for each 
subsampling and number of samples. Average and compute standard deviation of
loss values, and plot the results on a bar plot to reproduce the figure.
\begin{verbatim}
$ python subsample_random.py --path ./data --target drag \
  -ns [540|1080|2160] --dtype interpolated --window 3 \
  --noseed
$ python subsample_maxent.py --path ./data --target drag \
  -ns [540|1080|2160] --dtype interpolated --window 3 \
  --noseed
\end{verbatim}
Training command example:
\begin{verbatim}
$ python train-tf.py --epochs 100 --batch 2 --patience 12 \
  --test_frac 0.05 --arch lstm --window 3
\end{verbatim}

\noindent \textit{Expected Output:}

Bar plots comparing model accuracy and reproducibility. MaxEnt should yield lower training losses and standard deviations than random sampling. 

\noindent \textit{Correlation to Contributions:} 

Supports $C_2$ by demonstrating improved accuracy using MaxEnt.

\vspace{0.5em}

\balance

\noindent \textbf{Figure 7: Energy and accuracy (SST)}

\noindent \textit{Reproduction Instructions:}

\noindent Here is the command to perform \textbf{sampling} studies for SST-P1F100:

\begin{verbatim}
$ srun -n 32 python -u subsample.py \
  --path /path/to/P1F100Gn0050 \
  --dtype sst-binary --hypercubes [maxent|random] \
  --method [maxent|uips|full] --num_hypercubes 12 \
  --num_clusters 5 --num_samples 16384 \
  --nxskip 4 --nyskip 4 --nzskip 4 --nx 4098 \
  --ny 1024 --nz 4086 --nxsl 32 --nysl 32 --nzsl 32 \
  --gravity y --input_vars rhoy \
  --output_vars ee --cluster_var rhoy \
  >& subsample.out 
\end{verbatim}

\noindent Here is the configuration for sampling SST-P1F4:

\begin{verbatim}
$ srun -n 32 python subsample.py \
  --path /path/to/P1F4R32 \
  --dtype sst-binary --hypercubes [maxent|random] \
  --method [maxent|uips|full] --num_hypercubes 12 \
  --num_clusters 20 --num_samples 3277 \
  --nx 514 --ny 512 --nz 256 \
  --nxsl 32 --nysl 32 --nzsl 32 \
  --gravity y --input_vars u v w r \
  --output_vars p pv --cluster_var p pv \
  >& train.out
\end{verbatim}

\noindent For \textbf{training} for either case, use the following:

\begin{verbatim}
$ python train.py --epochs 1000 --batch 16 \
  --target p_full --window [1|2] --sequence [false|true] \
  --arch [MLP_Transformer|CNN_Transformer] >& train.out
\end{verbatim}

\noindent Notes:  
\begin{itemize}
\item When \texttt{{-}-method full} use \texttt{{-}-arch CNN\_Transformer}, otherwise use \texttt{MLP\_Transformer}. 
\item When \texttt{{-}-window 1} use \texttt{{-}-sequence false}, 
otherwise set to \texttt{true}. 
\item On Frontier, access can be made available to the yaml config files and slurm scripts used
to automate the launching of sampling and training runs for each configuration. 
In this case, one may simply perform the study as follows. In this case, use the 
provided \texttt{slurm.sh} scripts located in the \path{contrib/configs/SST/P1} and 
\path{contrib/config/SST/P1F100Gn0050} 
directory to run subsampling and training jobs on Frontier. 
Submit the job using:
\begin{verbatim}
$ sbatch slurm.sh
\end{verbatim}
\item This will run the following two commands for each of the YAML config files listed.
\begin{verbatim}
$ srun -n 32 python -u subsample.py $CASE.yaml
$ srun --ntasks-per-node=8 python -u train.py \ 
  $CASE.yaml
\end{verbatim}
\end{itemize}

\noindent After completion, extract \textbf{energy consumption} values using:

\begin{verbatim}
$ grep CPU.Energy subsample*.out
$ grep Total.Energy.Consumed train*.out
\end{verbatim}

\noindent Extract \textbf{loss metrics} using:
\begin{verbatim}
$ grep Evaluation train*.out
\end{verbatim}

Add CPU energy from subsampling to total energy from training to compute total energy cost.
Plot in Excel against loss values extracted from training outputs. 

\noindent \textit{Expected Output:}

Accuracy vs. energy plot showing that MaxEnt typically gives lower loss values at significantly lower energy costs. For SST-P1F4, we observed a 38X reduction by MaxEnt sampling just 10\% of the points as compared with training on fully dense hypercubes. We observed similar behavior with SST-P1F100, however, while the full case consumed 15\% more power, it ran faster than some of the subsampled cases, and gave a loss value above one, which is why the point does not show up on the plot. 

\noindent \textit{Correlation to Contributions:} 

Supports $C_3$ showing energy efficiency vs accuracy.

\vspace{0.5em}

\noindent \textbf{Figure 8: Scalability (SST)}

\noindent \textit{Reproduction Instructions:}

To perform the scalability study, we use the sample sampling commands that were used for Fig. 7, but
we varying the number of MPI processes from 1 to 512 in powers of two. For most studies, we
allocated eight nodes; for 512-process runs, 16 nodes were used. With the 
\begin{verbatim}
$ time srun -n [1,2,4,8,...,512] python subsample.py \
  configs/SST/P1/Hmaxent-Xmaxent.yaml
$ time srun -n [1,2,4,8,...,512] python subsample.py \
  configs/SST/P1F100Gn0050/Hmaxent-Xmaxent.yaml
\end{verbatim}
To compute the speedup we divide the times from the time command by the time it takes to complete a single process. 
The efficiency is the speedup value divided by the $N$ number of processes. Then, we plot these on a scatter plot in Excel, setting the x-axis to $log_2$ scale. 

\noindent \textit{Expected Output:}

Speedup and efficiency plots. SST-P1F100 shows quasilinear speedup up to 64 MPI processes, after which it falls down to about 10 at 512 MPI processes. SST-P1F4 shows sublinear scaling performance, reaching max speedup of 9 at 32 MPI processes. 

\noindent \textit{Correlation to Contributions:} 

Validates $C_4$ demonstrating scalability, albeit still could use improvements.

\end{document}